\documentclass[journal]{IEEEtran}
\usepackage{amsmath, amssymb, amsfonts, bm, bbm}
\usepackage{algorithmic}
\usepackage{algorithm}
\usepackage{array}
\usepackage[caption=false,font=normalsize,labelfont=sf,textfont=sf]{subfig}
\usepackage{textcomp}
\usepackage{stfloats}
\usepackage{url}
\usepackage{verbatim}
\usepackage{graphicx}
\usepackage{cite}

\usepackage{xspace}
\usepackage{booktabs}
\usepackage{placeins}
\usepackage{multirow}
\usepackage{pifont}
\usepackage{threeparttable}
\usepackage{color, soul}

\def\OurMethod{{DepthCropSeg++}\xspace}
\def\eg{\textit{e.g.}\xspace}

\begin{document}

\title{DepthCropSeg++: Scaling a Crop Segmentation Foundation Model With Depth-Labeled Data}


\author{
Jiafei~Zhang,
Songliang~Cao,
Binghui~Xu,
Yanan~Li, 
Weiwei~Jia,\\
Tingting~Wu,
Hao~Lu,~\IEEEmembership{Member,~IEEE},
Weijuan~Hu,
and~Zhiguo~Han
\thanks{This work was jointly supported by the Key Technology Research and Development Program of Shanghai under Grant No. 25N32800100, National Key R \& D Program of China under Grant No. 2023YFF1001502, National Natural Science Foundation of China Under Grant No. 32370435, Central Government's Guidance Fund for Local Science and Technology Development Under Grant No. 2024ZY-CGZY-19, Hubei Provincial Natural Science Foundation of China under Grant No. 2024AFB566, and PhenoTrait Foundation. Part of work was done when Binghui Xu was with School of Artificial Intelligence and Automation, Huazhong University of Science and Technology, Wuhan, 430074, China and when Binghui Xu and Songliang Cao were interns at PhenoTrait Technology. (Jiafei Zhang and Binghui Xu contributed equally to this work.) (Corresponding author: Weijuan Hu and Zhiguo Han.)}
\thanks{J. Zhang and Z. Han are with MetaPheno Laboratory, Shanghai 201114, China, and with PhenoTrait Technology Co., Ltd., Beijing 100096, China (e-mail: \{joyce.zhang,david.han\}@phenotrait.com).}

\thanks{B. Xu is with Wuhan Digital Engineering Institute, Wuhan 430205, China (e-mail: \{xubh810@foxmail.com).}

\thanks{S. Cao and H. Lu are with National Key Laboratory of Multispectral Information Intelligent Processing Technology, School of Artificial Intelligence and Automation, Huazhong University of Science and Technology, Wuhan, 430074, China (e-mail: \{songliangcao@126.com, hlu@hust.edu.cn).}

\thanks{Y. Li is with Hubei Key Laboratory of Intelligent Robot, School of Computer Science and Engineering, School of Artificial Intelligence, Wuhan Institute of Technology, Wuhan, 450205, China (e-mail: yananli@wit.edu.cn).}

\thanks{W. Jia is with Beijing Agricultural Technology Extension Station, Beijing 100029, China (e-mail: jiaweiwei2003@163.com).}

\thanks{T. Wu is with College of Mechanical and Electronic Engineering, Northwest A\&F University, Yangling 712100, China (e-mail: tt\_wu@nwsuaf.edu.cn).}

\thanks{W. Hu is with Laboratory of Advanced Breeding Technologies, Institute of Genetics and Developmental Biology, Chinese Academy of Sciences, Beijing 100101, China (e-mail: wjhu@genetics.ac.cn).}

}





\markboth{Journal of \LaTeX\ Class Files,~Vol.~14, No.~8, August~2021}%
{Shell \MakeLowercase{\textit{et al.}}: A Sample Article Using IEEEtran.cls for IEEE Journals}


\maketitle

\begin{abstract}
We introduce \OurMethod, a foundation model for crop segmentation, capable of 
segmenting different crop 
species under open in-field environment. 
Crop segmentation is a fundamental task for modern agriculture, 
which 
closely relates to many downstream tasks such as plant phenotyping, density estimation, and weed control. 
In the era of foundation models, a number of generic large language and vision models have been developed. These models have demonstrated remarkable real-world generalization due to significant model capacity and large-scale datasets.
However, current crop segmentation models mostly learn from 
limited data due to expensive pixel-level labelling cost, 
often performing well only under specific crop types or controlled environment. 
In this work, we follow the vein of 
our previous work DepthCropSeg, an almost unsupervised approach to crop segmentation, to scale up 
a cross-species and cross-scene crop segmentation dataset, with $28,406$ images across $30$+ species and $15$ environmental conditions. We also build upon a state-of-the-art semantic segmentation architecture ViT-Adapter architecture, enhance it with dynamic upsampling for improved detail awareness, and train the model with a two-stage self-training pipeline. To systematically 
validate model performance, we conduct comprehensive 
experiments to justify the effectiveness and generalization capabilities across multiple crop datasets.
Results demonstrate that \OurMethod achieves $93.11\%$ mIoU on 
a comprehensive testing set, outperforming both supervised baselines and general-purpose vision foundation models like Segmentation Anything Model (SAM) by significant margins ($+0.36\%$ and +$48.57\%$ respectively). The model particularly excels in challenging scenarios including night-time environment ($86.90\%$ mIoU), high-density canopies ($99.86\%$ mIoU), and unseen crop varieties ($90.09\%$ mIoU), 
indicating a new state of the art for crop segmentation.
\end{abstract}

\section{Introduction}
\IEEEPARstart{P}{lant} phenotyping refers to the comprehensive measurement and analysis of plant traits, including morphology, physiology, and biochemical characteristics~\cite{HAN2025}. Accurate phenotyping is essential for understanding plant responses to environmental changes and for developing strategies to enhance plant resilience and productivity. Crop segmentation is one of fundamental tasks in this regime, 
    as it provides the basis for extracting various traits from plant images.
    %
    In quantitative remote sensing of vegetation, crop segmentation also supports the extraction of canopy structure parameters such as Green fraction (GF), Green area index (GAI), and Leaf angle distribution (LAD), which dynamically affects the productivity of crops~\cite{gao2024bridging}.
    The accuracy of crop segmentation directly influences the performance of 
    plant phenotyping~\cite{XU2025110337} and other related agricultural vision tasks such as density estimation~\cite{sheather2004density,scott2015multivariate}, cover crop identification~\cite{cruz2012multi} and weed control~\cite{gao2024accurate}. 

    To enhance performance in these agricultural downstream tasks, researchers have explored many methods~\cite{hamuda2016survey}. Currently, the mainstream ones are those based on deep learning. However, existing crop segmentation models~\cite{ZHANG2024108740,10621585, ebrahimi2025helps} generally exhibit limited specificity, often achieving optimal performance only under specific crop varieties and imaging conditions. 
    This is because the inherent particularities of agricultural imagery primarily constrain the generalization capabilities of segmentation models: i) there are significant morphological and textural variations between different varieties of crops; ii) significant variations exist in the same plant across different growth stages~\cite{luo2024semantic}; iii) complex field environments present intense illumination variations, cluttered backgrounds, and inconsistent imaging conditions; iv) if crops are densely planted, severe shading may also occur~\cite{Xiong2019}; and v) the high cost of acquiring high-quality agricultural images with fine-grained annotations limits the construction of large-scale datasets~\cite{zamani2025novel}. 
    Faced with these challenges, current research usually aims to improve performance only by modifying model architectures, yet they are often limited to small single-category datasets. Consequently, When application scenarios change, it becomes necessary to recollect data and retrain models. Due to the typical crop growth cycle lasting from three months to ten months, data collection is often challenging. Meanwhile, there will be a large amount of repetitive pixel-level annotation work.
    To improve the efficiency of related tasks, building a 
    crop segmentation model with 
    cross-species and cross-scene adaptability has become an urgent necessity.

   In the field of computer vision, foundation model is the key to improve model performance in various downstream tasks.
   In recent years, general purpose vision foundation models (VFMs), exemplified by SAM~\cite{kirillov2023SAM}, DINOv2~\cite{Oquab2024} and InternVL~\cite{chen2024internvl}, have achieved significant advancements. 
   These models demonstrate the potential for enhancing the generalization of the model through large-scale, heterogeneous data training. However, due to the difference in distribution between natural images and agricultural images, performance tends to decline when applied to downstream agricultural tasks. The solution to this challenge is straightforward: collect large-scale agricultural data and train a foundation model on it. 
   To bypass the high cost of pixel-level annotation, we previously proposed DepthCropSeg~\cite{cao2025blessing}, an almost unsupervised segmentation method that generates high-quality pseudo-labels from depth maps without manual pixel-level annotations. The proposed DepthCropSeg facilitates comprehensive exploitation of massive unlabeled agricultural imagery.

    \begin{figure*}[!t]
        \centering
        \includegraphics[width=\textwidth]{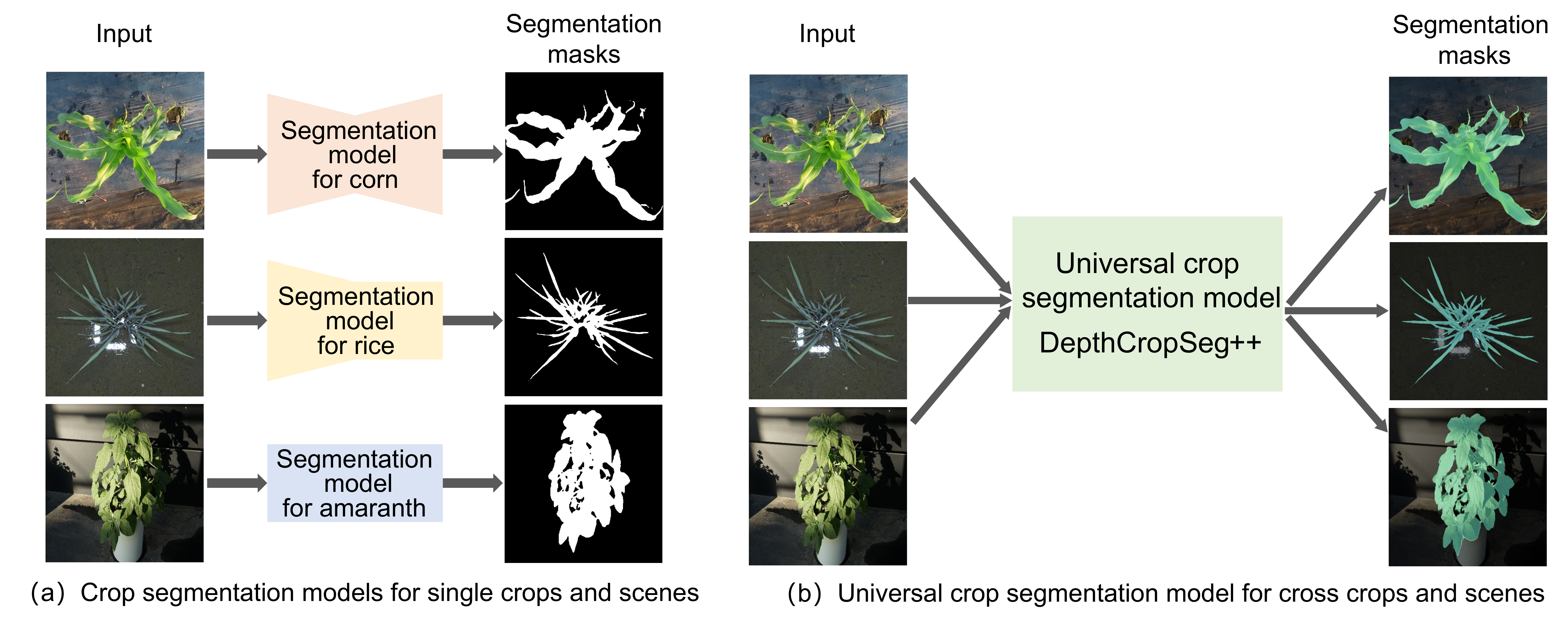}
        \caption{Comparison between specialized and general-purpose crop segmentation models. (a) Specialized models require separate training for each crop variety, while (b) the general-purpose model enables precise segmentation of arbitrary plant species and scenarios using a single strongly generalized model}\label{fig:specific_general_model}
    \end{figure*}
    
    Building upon this foundation, we present an enhanced version \OurMethod, a foundation model for crop segmentation. As shown in Fig.~\ref{fig:specific_general_model}, compared with traditional specialized models, our proposed framework adopts a unified architecture that enables cross-crop and cross-environment image segmentation, which demonstrating superior adaptability and generalization capabilities. First, we utilize DepthCropSeg for rapid pseudo-masks generation and expand the training dataset from $3,577$ to $28,406$ images, an order of magnitude greater than the number of images used to train DepthCropSeg. The large-scale dataset includes over 30 plant species, including major crops like rice, corn, wheat, and soybean. The images were captured under diverse environmental conditions (e.g., sunny, cloudy, nighttime) and from multiple viewing angles. To improve the model’s ability to handle complex scenes, we specifically augmented the dataset with images densely populated with crops, as these are  challenging for DepthCropSeg. 
    Second, we integrate the dynamic upsampling operator FADE(Fusing the Assets of Decoder and Encoder for Task-Agnostic Upsampling)~\cite{lu2024fade} into ViT-Adapter. Unlike general-purpose applications, agricultural images often involve intricate object shapes, overlapping leaves, and ambiguous boundaries. These segmentation boundaries directly affect the calculation of phenotypic parameters. 
    By fusing the assets of decoder and encoder, FADE can upsample feature maps accurately, which enhance the ability of model to solve fine structures.
    Finally, we design a two-stage training strategy to progressively enhance segmentation quality. In the first stage, we train a base model using the initial pseudo-labels generated by DepthCropSeg. 
    During the second stage, predictions from the base model are leveraged to refine pseudo-labels by preserving consistent regions. Inconsistent regions are excluded from loss calculation. This progressive refinement not only mitigates the noise inherent in unsupervised labels but also significantly boosts the model’s generalization across unseen crop types and field conditions. 
    To systematically evaluate model performance, we conduct comprehensive validation of the segmentation effectiveness and generalization capabilities of \OurMethod across multiple crop datasets. Experimental results demonstrate that our method not only achieves superior performance on known crops but also exhibits robust generalization to novel varieties and complex environments, outperforming the upper bound of models trained with manual annotations.

    The contributions of this work include the following:

    \begin{itemize}
        \item \OurMethod: a foundation model for crop segmentation with cross-species and cross-scene capabilities;
        
        \item We scale up a large-scale crop segmentation dataset following depth-informed pseudo labelling and fast manual screening, including $28,406$ images, $30+$ species, and $15$ environmental conditions;
        
        \item We upgrade the base model architecture of DepthCropSeg to the state-of-the-art ViT-Adapter and further improve its detail awareness with dynamic upsampling.
    \end{itemize}

\section{Datasets and Methods}\label{sec:method}
In this section, we first describe the baseline model employed in \OurMethod, and review the process by which DepthCropSeg generates pseudo masks from depth information. Subsequently, we introduce how the cross-scenario crop dataset is extended for training purposes. Finally, we present the training details and strategies of our foundational crop segmentation model, \OurMethod.

\begin{figure}[!t]
    \centering
    \includegraphics[width=\columnwidth]{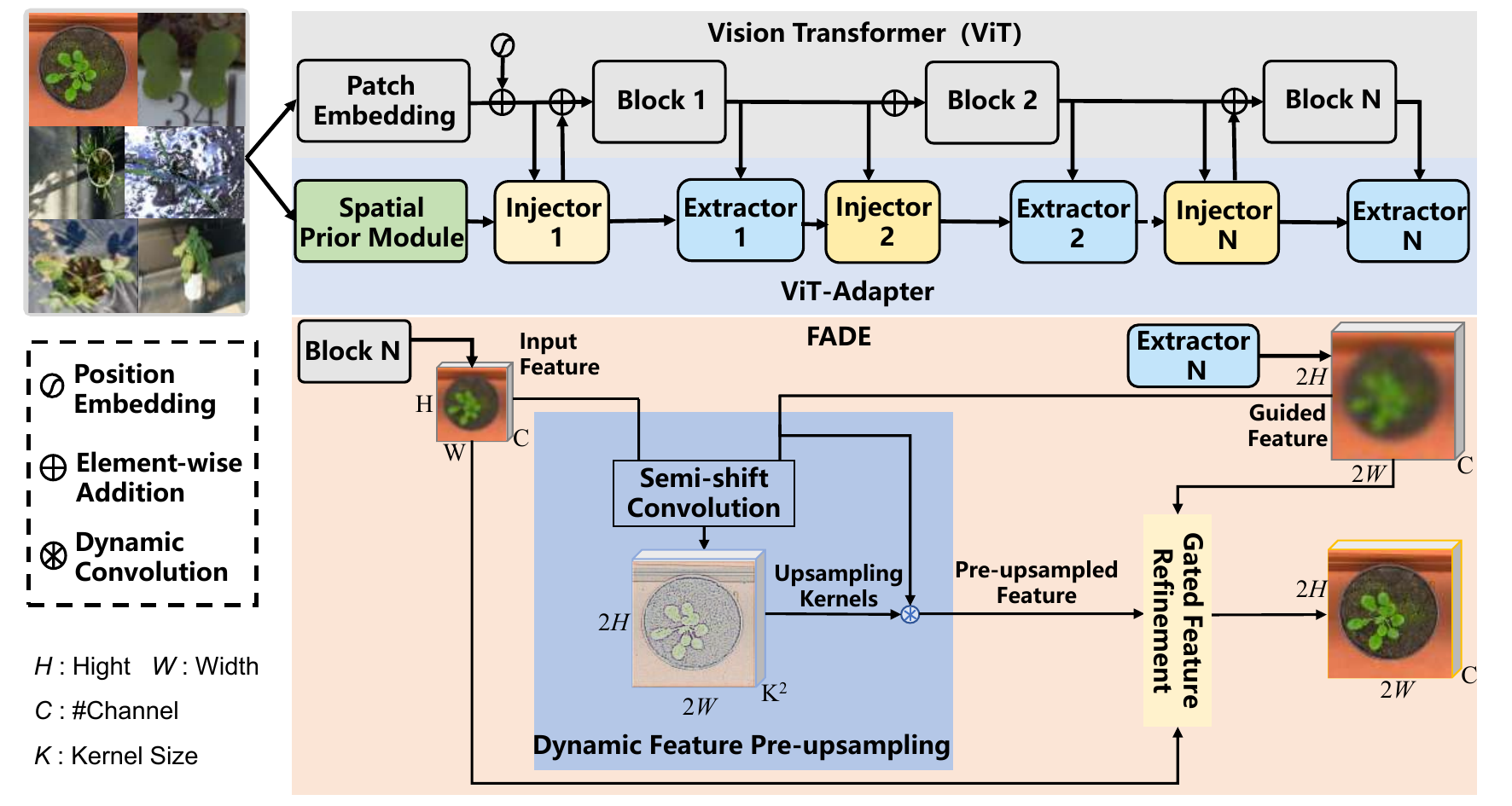}
    \caption{Architecture of the baseline model in \OurMethod: ViT-Adapter with FADE upsampling.}\label{fig:Architecture}
\end{figure}

\subsection{Baseline: Vision Transformer with ViT-Adapter}\label{sec:ViT-Adapter}
In recent years, attention-based Transformer models have achieved remarkable advancements in computer vision tasks. Compared with traditional convolutional neural networks, Transformers demonstrate superior global feature modeling capabilities, making them particularly well-suited for object recognition and segmentation tasks in complex scenarios. Consequently, designing Transformer-based visual backbone networks has become a prominent research direction in computer vision. ViT-Adapter constitutes an architectural advancement, featuring a task-specific dense prediction adapter tailored for the original Vision Transformer (ViT). In this study, the ViT-Adapter is adopted as the core architecture to increase the generalizability of crop segmentation models in diverse crop varieties, illumination conditions, and field settings.
    
    Current dominant technical approach for vision Transformers typically follows a three-stage pipeline: architectural design of pyramidal/hierarchical structures, followed by large-scale pre-training on image datasets, and culminating in downstream task adaptation through fine-tuning. 
    As illustrated in Fig.~\ref{fig:Architecture}, ViT-Adapter demonstrates enhanced flexibility through its architectural design. The framework employs a straight-tube structured general model (e.g., the original ViT), which can leverage multimodal data beyond images during pre-training. When adapting to downstream tasks, ViT-Adapter introduces image-related prior knowledge (inductive biases) into the pre-trained ViT network through randomly initialized adapter modules, making the model more aligned with these target tasks. This approach achieves decoupling between upstream pre-training and downstream fine-tuning phases while delivering performance comparable to or even surpassing pyramidal/hierarchical models.

    ViT-Adapter was the first Transformer-based segmentation model to surpass $60\%$ mIoU on the ADE20K benchmark, later achieving up to $61.5\%$ mIoU. Without using external datasets such as Objects365, it also reached state-of-the-art performance in object detection and instance segmentation. Additionally, compared to the original ViT, ViT-Adapter reduces computational complexity by introducing lightweight adapter modules, which makes it more suitable for real-world agricultural applications with limited computational resources. This ViT-Adapter baseline forms the foundation upon which we introduce further enhancements for robust crop segmentation.

    \begin{table*}[!t] \small
        \centering
        \caption{Summary of the supervised benchmark training set}
        \label{tab:11406}
        \renewcommand{\arraystretch}{1.2}
        \begin{tabular}{@{}lccccccc@{}}
            \toprule
            Dataset & CWFID & CVPPP & EWS & PhenoBench & VegAnn & Crop And Weed & Total \\ \midrule
            \#Images & 40 & 810 & 164 & 1407 & 3289 & 5694 & 11406 \\
            \bottomrule
        \end{tabular}
    \end{table*}

    \begin{figure*}[!t]
        \centering
        \includegraphics[width=\textwidth]{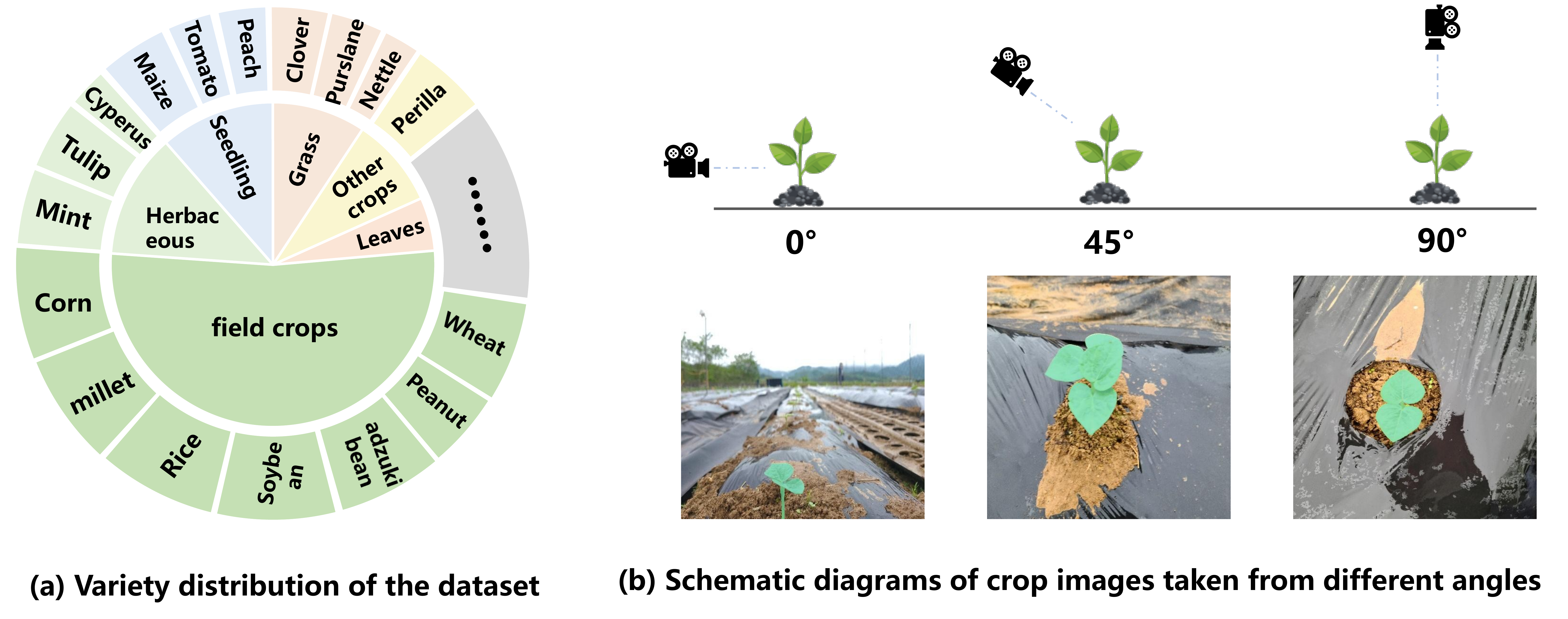}
        \caption{Species distribution and imaging angles in extended datasets}\label{fig:Category}
    \end{figure*}

\subsection{Optimization: FADE for Boundary-Aware Upsampling}\label{sec:FADE}
Crop segmentation represents a dense prediction task that requires high-quality feature upsampling to recover high-resolution features from input images. However, the bilinear interpolation used in the upsampling of ViT-Adapter introduces smoothing effects, leading to blurred crop boundaries and detail loss, thereby compromising model adaptability in complex field environments.

To address this issue, we replace the upsampling layers in the ViT-Adapter decoder with a task-agnostic dynamic upsampling operator called FADE (Fusing the Assets of Decoder and Encoder). As shown in Fig.~\ref{fig:Architecture}, FADE integrates both shallow CNN encoder features and global Transformer decoder features to dynamically generate upsampling kernels. This allows more precise recovery of crop boundaries and improves segmentation accuracy for fine structures such as leaves and stems.

Unlike previous methods such as CARAFE~\cite{wang2019carafe} or IndexNet~\cite{lu2019indices}, which tend to perform well only on specific tasks, FADE offers strong generalization across region-sensitive and detail-sensitive tasks. It achieves this by computing gating weights that adaptively fuse encoder and decoder features. Near object boundaries, FADE assigns higher weights to encoder features to sharpen edges, while inside object regions, decoder features dominate to ensure semantic consistency. Moreover, FADE is computationally efficient and does not rely on high-resolution guidance features, making it especially suitable for deployment on resource-constrained agricultural platforms such as UAVs and field robots. The integration of FADE thus represents the first key component in our optimization pipeline for \OurMethod.

\subsection{DepthCropSeg Revisited}\label{sec:DepthCropSeg}
To reduce the annotation burden in crop segmentation, DepthCropSeg~\cite{cao2025blessing} introduces an almost unsupervised pipeline that directly generates pseudo masks from monocular depth maps, avoiding the need for pixel-level manual labels.

At the core of this approach is Depth Anything V2~\cite{depth_anything_v2}, a high-performance monocular depth estimation model capable of producing high-resolution, semantically meaningful depth maps from single RGB images. Given an input crop image, Depth Anything V2 infers a corresponding depth map that captures spatial structure and depth discontinuities, which is critical for distinguishing foreground plants from the background.

To transform the raw depth map into a pseudo mask, DepthCropSeg proposes a gradient-guided histogram thresholding method. This process begins with depth normalization, followed by edge detection, gradient-weighted enhancement, and sigmoid-based curve fitting. The final threshold is selected based on the point of maximum gradient change in the fitted curve, enabling robust separation between crop and background. This multi-stage process is designed to be adaptive across different illumination conditions and crop varieties. It is crucial to note that the key to accurate mask generation lies in the relative depth gap between the plant and the ground, regardless of the absolute depth values. Additionally, thanks to the robust generalization capabilities of Depth Anything V2, this method can accurately extract plant masks in most complex scenarios (\eg, water reflections, shadows).

To further improve pseudo label quality, a coarse-to-fine manual filtering step is employed. Masks with significant noise or structural inaccuracies are removed by lightweight manual intervention enables the efficient generation of a large number of high-quality pseudo masks suitable for model training.

DepthCropSeg also incorporates a depth-informed two-stage self-training framework. In the first stage, a segmentation model is trained using the selected pseudo masks. This model is then used to infer segmentation predictions on the same dataset. A trimap is constructed by comparing the original pseudo masks and predicted masks, defining confident foreground, confident background, and uncertain regions. In the second stage, the model is fine-tuned using this refined supervision, where ambiguous pixels are excluded from the loss computation. Optional depth-guided post-processing further enhances mask precision.

In this work, we build upon and extend DepthCropSeg by leveraging its ability to rapidly generate high-quality pseudo crop masks from depth images. Specifically, we utilize this mechanism to construct an expanded large-scale crop segmentation dataset, which serves as the foundation for training our baseline segmentation model \OurMethod.

\newcolumntype{C}[1]{>{\centering\arraybackslash}p{#1}} 
    \begin{table*}[!t]\small
        \centering
        \caption{Comparison of Extended Training Datasets}
        \label{tab:dataset_extend}
        \renewcommand{\arraystretch}{1.5} 
        \addtolength{\tabcolsep}{0pt}
        \begin{tabular}{@{}lC{1cm}C{1cm}C{2cm}C{2cm}C{1cm}C{2.8cm}@{}}
            \toprule
            Dataset & Quantity & Classes & Species & Resolution & Location & Acquisition Device \\ \midrule
            CWD30 & 219,778 & 30 & Crops/Weeds & 4032×3024 & Roadsides & Mixed \\
            Deep Weeds & 17,509 & 9 & Carrot/Weeds & 1296×966 & Fields & Tripod \\
            Plant Seedlings & 11,078 & 12 & Weeds & 355×355 & Trays & Tripod \\
            Apple Leaf & 1,730 & 12 & Weeds & 355×355 & Trays & Tripod \\
            PDD271 & 101,165 & 271 & Fruits/Vegetables & 256×256 & Fields & Handheld Camera \\
            Night Rice & 288 & 1 & Rice & 2318×2318 & Fields & Phenotyping Platform \\
            \bottomrule
        \end{tabular}
    \end{table*}

\subsection{Cross-Scenario Dataset Extension}
\label{sec:dataset_extend}

The performance of crop segmentation models largely depends on the quality and diversity of training data. A limited or homogeneous dataset may fail to capture the variability of real-world agricultural conditions, resulting in overfitting and poor generalization. To address this, we construct a comprehensive cross-scenario dataset for robust model training, consisting of both a fully supervised benchmark dataset and an expanded set of large-scale pseudo-labeled samples.

\subsubsection{Fully Supervised Benchmark Training Set}
Due to the high cost of pixel-level annotations in agricultural scenarios, the number of manually labeled crop segmentation datasets remains limited. We compile six publicly available datasets containing a total of 11,406 high-quality, human-annotated crop images. These datasets, primarily developed for specific tasks, are summarized in Table~\ref{tab:11406}. They cover a narrow range of crop types and lack sufficient environmental diversity (e.g., lighting variation, background clutter), limiting their applicability in diverse field conditions. Therefore, we use this dataset to establish an upper-bound performance for fully supervised training.

    \begin{figure}[!t]
        \centering
        \includegraphics[width=\columnwidth]{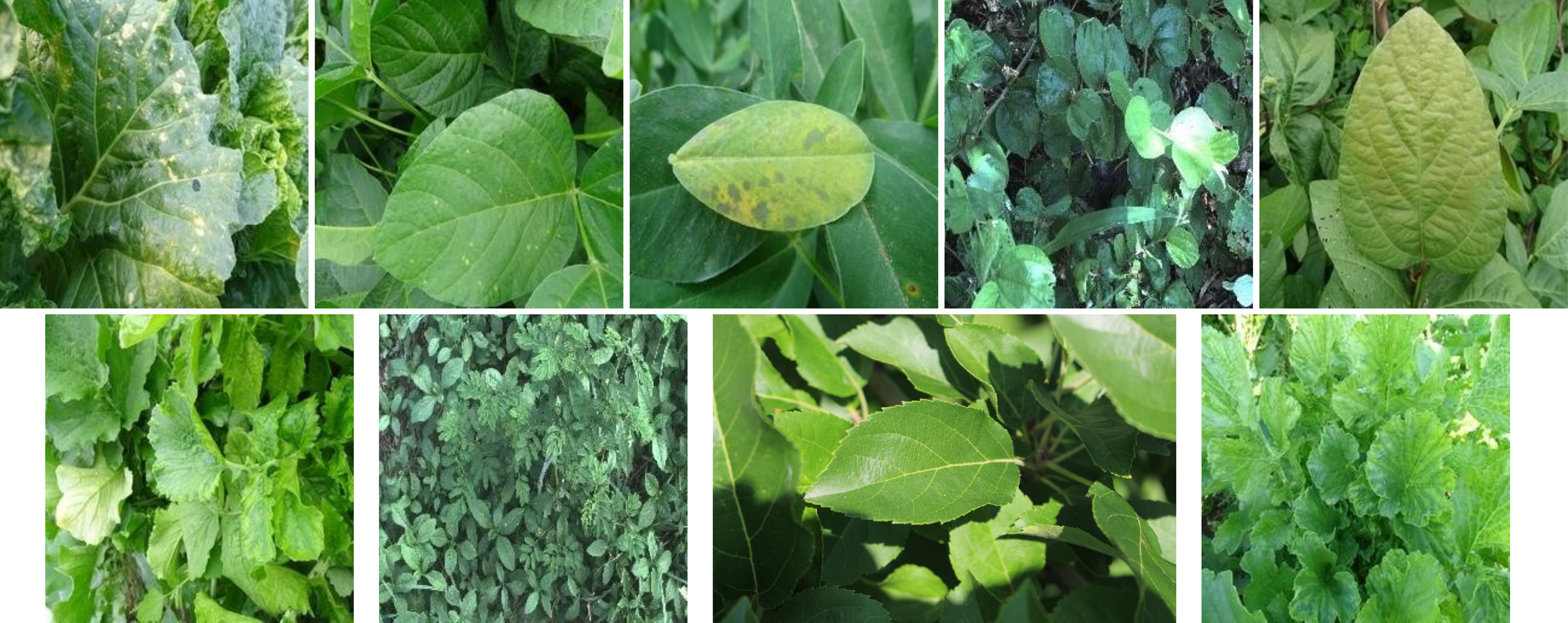}
        \caption{Examples of full-coverage crop images}\label{fig:full_coverage}
    \end{figure}
    
    \subsubsection{Large-scale Cross-Scenario Extended Dataset}

    To enhance the segmentation accuracy and generalizability of \OurMethod, we propose a depth-guided dataset expansion strategy that addresses three fundamental constraints: insufficient data scale, limited crop variety, and complex environmental conditions.
    Our expansion is driven by three key principles:

    \begin{enumerate}
        \item[(1)] \textbf{Cross-Species Expansion}. Incorporating 5 public datasets containing diverse crop species, with species distribution shown in Fig.~\ref{fig:Category}(a). This enhances model cross-species adaptability through morphological and textural variations. Since our task is crop binarization, incorporating more crop varieties will not introduce sample imbalance but instead serve to enhance model robustness.
        
        \item[(2)] \textbf{Multi-Environment Expansion}. New datasets include images captured under various conditions: different lighting (sunny, cloudy, night), imaging angles (Fig.~\ref{fig:Category}(b)), and field backgrounds. Notably, a self-collected nighttime rice dataset supplements the underrepresented nighttime scenes.
        
        \item[(3)] \textbf{Full-Coverage Image Expansion}. Fully vegetated images shown as Fig.~\ref{fig:full_coverage} are difficult to segment from DepthCropSeg. Thus, we add such images to train and test sets. Their pseudo-masks are generated as all-ones matrices matching image dimensions.
    \end{enumerate}

    \begin{figure}[!t]
        \centering
        \includegraphics[width=\columnwidth]{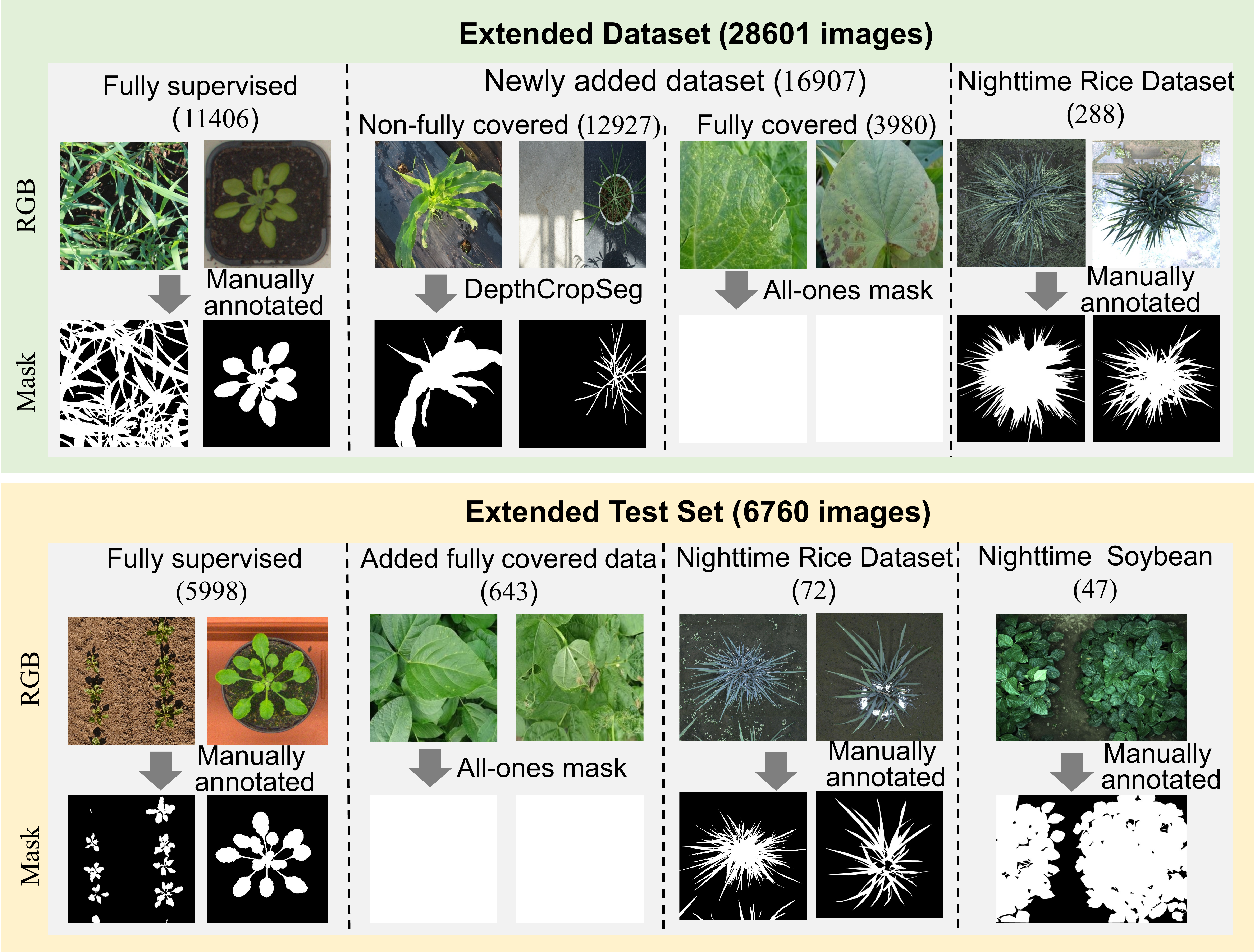} 
        \caption{Composition of extended training and test sets}
        \label{fig:Dataset_Sample}
    \end{figure}

Specifically, the extended training set comprises $5$ unlabeled public datasets and $1$ manually annotated night rice dataset, significantly expanding crop varieties, imaging angles, and lighting conditions. Table~\ref{tab:dataset_extend} compares these $6$ extended datasets:
    
    \begin{itemize}
        \item \textbf{CWD30~\cite{ilyas2025cwd30}}: Contains $219,770$ high-resolution images of $20$ weed and $10$ crop species across growth stages, angles, and environments from diverse geographical locations.
        
        \item \textbf{Plant Seedlings~\cite{v2plantseedlings}}: $5,539$ images of $12$ common Danish agricultural species (e.g., sugar beet, mustard) in 355×355 resolution, captured under controlled conditions.
        
        \item \textbf{Apple Leaf~\cite{appleleafdisease}}: $1,730$ images categorizing healthy/diseased apple leaves (scab, powdery mildew), captured via tripod-mounted cameras.
        
        \item \textbf{Deep Weeds~\cite{DeepWeeds2019}}: $17,509$ images of 8 Australian weed species in complex grassland environments.
        
        \item \textbf{PDD271~\cite{PDD271}}: $220,592$ images covering $271$ plant disease categories across fruits, vegetables, and field crops, captured $20$-$30$cm from plants.
        
        \item \textbf{Night Rice Dataset~\cite{XU2025110337}}: $360$ manually annotated images captured by high-throughput phenotyping platforms TraitDiscover (PhenoTrait Technology Co., Ltd.) located in various regions across China under controlled night lighting.
    \end{itemize}

    For unlabeled data, we partition images based on coverage completeness (fully covered vs. partially covered) and apply different pseudo-labeling strategies: 
    \begin{itemize}
        \item \textbf{Non-full-coverage images}: we applied the DepthCropSeg pipeline to generate pseudo masks and manually filtered $12,927$ high-quality samples from $260,260$ images. This process takes about $12$ hours.
        The quality control protocol follows our prior work, DepthCropSeg. Specifically, we exclude data instances that are fatally flawed or contain only partially labeled crops as foreground.
        
        \item \textbf{Full-coverage images}: We manually selected 4,623 samples (3,980 training, 643 testing) and assigned all-ones mask to each full-coverage image. The process of selecting full-coverage images takes $2$ hours.
    \end{itemize}
    
    \begin{figure*}[!t]
        \centering
        \includegraphics[width=\textwidth]{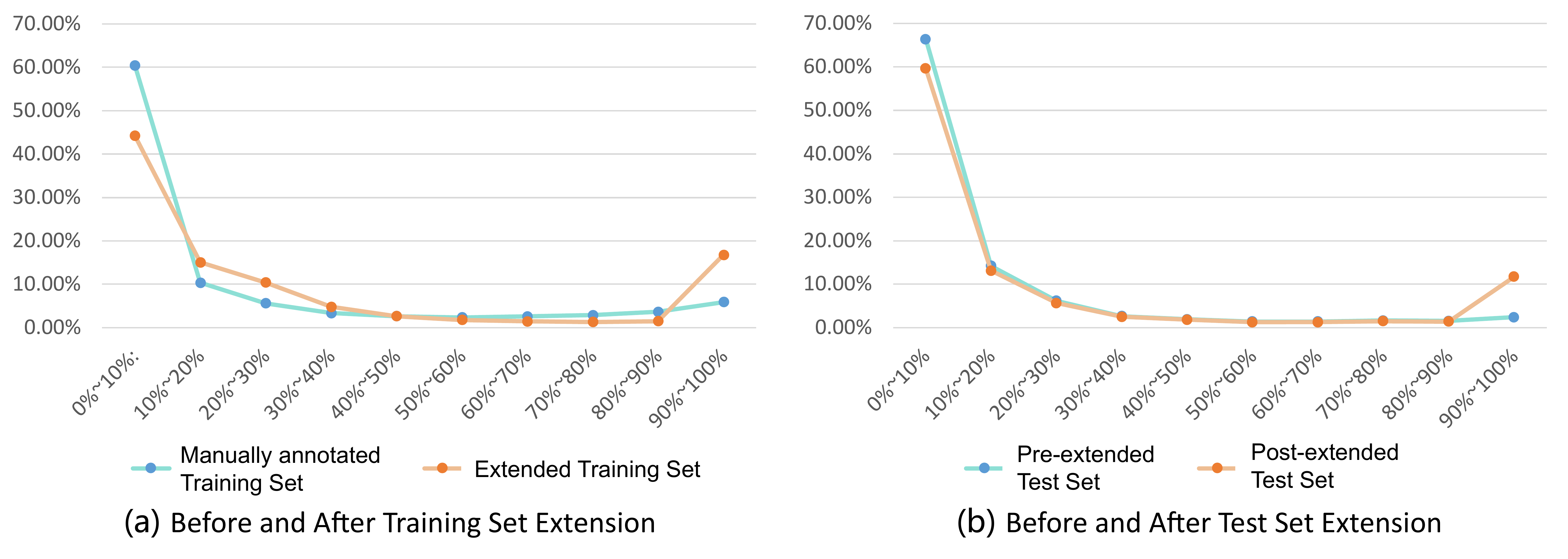} 
        \caption{Canopy coverage distribution before and after dataset expansion}
        \label{fig:canopy_distribution}
    \end{figure*}
    
    The final training set consists of $28,601$ images, combining $11,406$ supervised samples, 
    $17,195$ pseudo-labeled samples generated via depth estimation and manual filtering.
    The new test set comprises $6,760$ images from $10$ datasets, plus $72$ from nighttime rice and $47$ manually labeled nighttime soybean images for generalization assessment.
    
    Fig.~\ref{fig:Dataset_Sample} illustrates the composition of the dataset, with detailed statistics provided in Table~\ref{tab:extended_dataset}. Fig.~\ref{fig:canopy_distribution} shows increased high-density canopy coverage in extended datasets, better reflecting field conditions.

    \newcolumntype{D}{>{\centering\arraybackslash}p{2cm}}
    \begin{table*}[!t]\small
        \centering
        \caption{Statistics of Extended Training and Test Sets}
        \label{tab:extended_dataset}
        \renewcommand{\arraystretch}{1.2}
        \addtolength{\tabcolsep}{0pt}
        \begin{tabular}{@{}lDDDDDD@{}}
            \toprule
            \textbf{Dataset}  & \multicolumn{3}{c}{\textbf{Training Set}} & \multicolumn{2}{c}{\textbf{Test Set}} \\ \cmidrule(lr){2-4} \cmidrule(lr){5-6} 
            & {DepthCropSeg} & {Full-Coverage} & {Total} & {Full-Coverage} & {Total}  \\ \midrule
            Supervised Set  & - & - & 11,406 & - & 5,998 \\
            CWD30        & 7,115 & - & 7,115 & - & - \\
            Apple Leaf  & 390 & 278 & 668 & 30 & 30 \\
            Deep Weeds & 196 & 802 & 998 & 150 & 150 \\
            Plant Seedlings  & 4,429 & - & 4,429 & - & - \\
            Sample   & 797 & 2,900 & 3,697 & 463 & 463 \\
            Night Rice  & - & - & 288 & - & 72 \\ 
            Night Soybean  & - & - & - & - & 47 \\ \midrule
            Total  & & & 28,601 & & 6,760 \\
            \bottomrule
        \end{tabular}
    \end{table*}

    \subsection{Training Strategies and Implementation Details}\label{sec:strategy}

    This section presents the proposed \OurMethod training pipeline, including data pre-processing, training configurations, and our two-stage training strategy.

    \subsubsection{Data Preprocessing}   

    To improve model robustness under diverse field conditions, we apply a set of data augmentation techniques tailored to crop segmentation tasks:

    \begin{itemize}
        \item[(1)] \textbf{Geometric transformations:} Random scaling (to $3584 \times 896$), cropping, and horizontal flipping (50\% probability) simulate plant size variations, camera angles, and spatial orientation, enhancing the model's generalization to different crop structures and growth stages.
        
        \item[(2)] \textbf{Color perturbations:} To address illumination variability (e.g., sunny, cloudy, nighttime conditions), random adjustments in brightness, contrast, saturation, and hue are applied to reduce dependency on specific lighting.
    
        \item[(3)] \textbf{Normalization:} All images are normalized with their dataset-specific mean and standard deviation, thereby stabilizing training and mitigating cross-dataset domain shifts.
    
        \item[(4)] \textbf{Image padding:} To preserve small-object information during resizing and ensure consistent input dimensions, images are zero-padded when necessary.
    \end{itemize}

    These preprocessing techniques enable the model to better adapt to diverse crop types and field imaging scenarios while reducing overfitting.

    \begin{figure}[!t]
        \centering
        \includegraphics[width=\columnwidth]{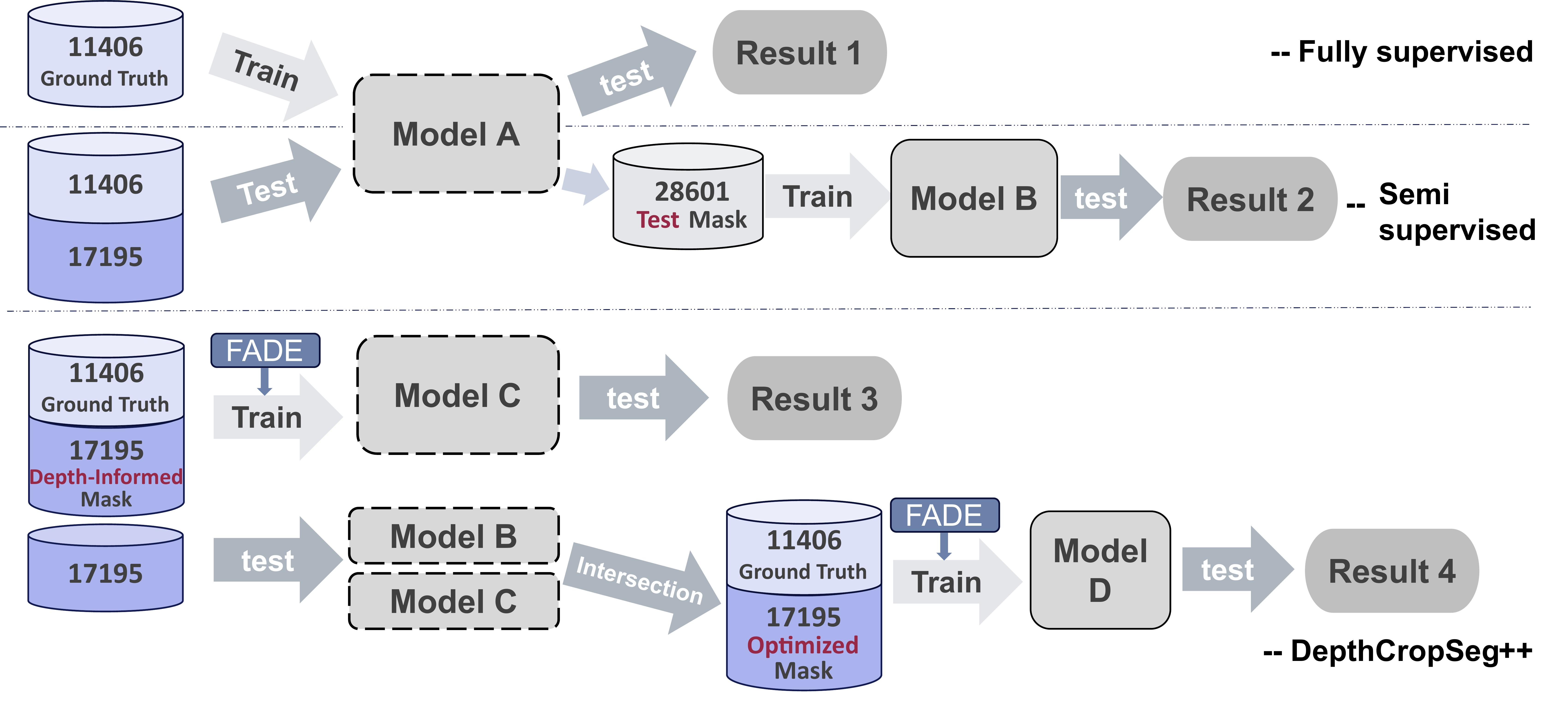}
        \caption{Two-stage training strategy for \OurMethod.}
        \label{fig:experiment_plan}
    \end{figure}

    \begin{figure*}[!t]
        \centering
        \includegraphics[width=\textwidth]{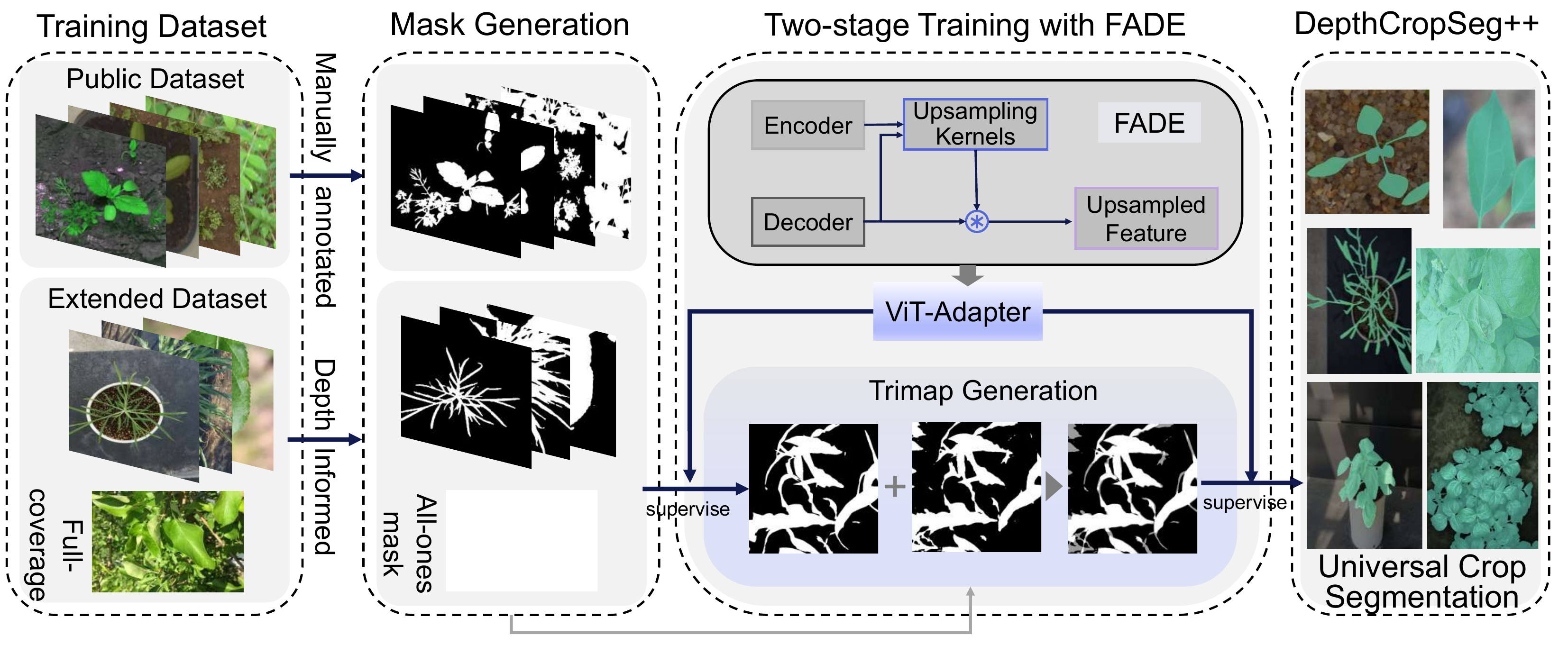}
        \caption{Overview of the \OurMethod technical pipeline.}
        \label{fig:CropSegNet_pipeline}
    \end{figure*}

    \subsubsection{Implementation and Training Details}\label{subsec:details}
    
    \paragraph{Training Setting} 
    We adopt the BEiT-Adapter-Large architecture as the backbone of \OurMethod, enhanced with multi-scale deformable attention modules for effective feature representation. The model is initialized with weights pretrained on the COCO-Stuff 164k dataset for 80k iterations. Fine-tuning on our agricultural dataset is performed using the AdamW optimizer with a learning rate of $2e-5$, weight decay of $0.05$, and a layer-wise learning rate decay factor of $0.9$.

    \paragraph{Experimental Platform} All experiments are conducted on the Ubuntu 20.04 system with $\tt Python$ version $3.8$, $\tt PyTorch$ version $1.9.0$, CUDA version $11.8$, and on a workstation with four $48$ GB RTX$A6000$ GPUs, two $10$-core Intel Xeon Silver 4210R CPUs, and $256$ GB RAM.

    \subsubsection{Two-Stage Self-Training Strategy}
    
    As illustrated in Fig.~\ref{fig:experiment_plan}, we adopt a two-stage self-training framework to maximize the utility of the expanded dataset (Table~\ref{tab:extended_dataset}). The full training set contains $28,601$ images, including $11,406$ fully labeled samples and $17,195$ pseudo-labeled samples generated via depth estimation and manual filtering.
    
    We first train \OurMethod on the complete training set using the current pseudo masks. This stage enables the model to learn coarse semantic priors from both manually labeled and pseudo labeled images.
    
    The trained model is used to re-infer segmentation maps on the entire training set. We then compare the new predictions with the initial pseudo masks and construct a trimap: pixels with consistent predictions are retained, while uncertain pixels (i.e., inconsistent labels) are assigned a value of 255 and ignored during training. This filtered supervision is used to retrain the model in the second stage.
    
    In both stages, the ViT-Adapter decoder is enhanced by replacing bilinear upsampling with the FADE module, which improves boundary segmentation accuracy, particularly in densely vegetated and fine-structure regions. In the original ViT-Adapter, the low-resolution features produced by the ViT backbone blocks are upsampled and then added to the multi-scale features extracted by the adapter. The default upsampling method is bilinear interpolation, and we replace all such upsampling operations with the proposed FADE module. The pipeline of \OurMethod is shown in Fig.~\ref{fig:CropSegNet_pipeline}.

    \subsubsection{Testing Protocol}
    
    The evaluation is conducted on a $6,760-image$ test set covering various crops, imaging angles, and lighting conditions (Table~\ref{tab:extended_dataset}). Notably, the test set exclusively includes $47$ manually labeled nighttime soybean images. This exclusive inclusion is particularly for evaluating the model's generalization capability under both unseen and challenging field conditions.


    \section{Results}\label{sec:exp}
    This section presents the performance of \OurMethod through qualitative and quantitative comparisons with fully supervised benchmarks, semi-supervised baselines, and general vision foundation models (SAM and HQ-SAM) across various complex environments.

    \begin{figure*}[!t]
        \centering
        \includegraphics[width=\textwidth]{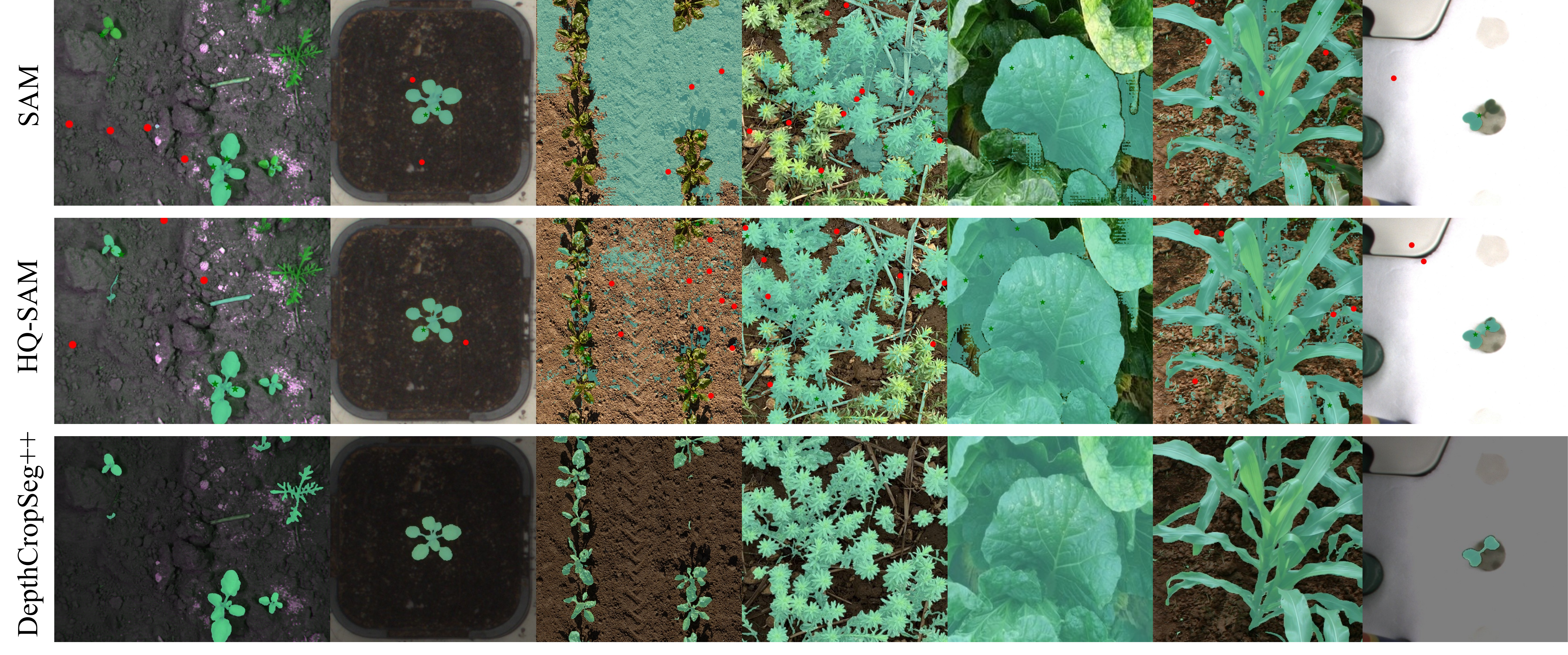} 
        \caption{Visual comparison with vision foundation models}\label{fig:SAM_compare}
    \end{figure*}

    \begin{table}[!t]
        \footnotesize
        \centering
        \caption{Benchmark Comparison Results of \OurMethod}
        \label{tab:benchmark_03}
        \renewcommand{\arraystretch}{1.5}
        \addtolength{\tabcolsep}{-1pt}
        \begin{tabular}{@{}ccccc@{}}
        \toprule
        \multirow{2}{2.2cm}{\centering Depth-based\\Pseudo-masks}  & \multirow{2}{2.2cm}{\centering Two-stage\\Self-training} & \multirow{2}{2.2cm}{\centering FADE\\Upsampling}  & \multirow{2}{*}{\centering mIoU} \\
        & & & \\
        \midrule
        & & & 92.36 \\
        \ding{51} & & & 92.86 \\
        \ding{51} & \ding{51}  & & 92.88 \\
        \ding{51} & \ding{51} & \ding{51} & 93.11 \\
        \midrule
        \multicolumn{3}{l}{SAM} &  32.80 \\
        \multicolumn{3}{l}{HQ-SAM} &  44.54 \\
        \multicolumn{3}{l}{GWFSS Top-Solution} & 69.50  \\
        \multicolumn{3}{l}{Fully Supervised} & 92.75 \\
        \bottomrule
        \end{tabular}
    \end{table}

    \begin{table}[!t]
        \footnotesize
        \centering
        \caption{Comparison results of different upsampling method}
        \label{tab:compare_upsample}
        \renewcommand{\arraystretch}{1.5}
        \addtolength{\tabcolsep}{-1pt}
        \begin{tabular}{@{}lcccccc@{}}
        \toprule
        \textbf{Method}  & \textbf{Params(M)} & \textbf{GFLOPs} & \textbf{FPS} & \textbf{Memory(G)}& \textbf{mIoU} \\
        \midrule
        bilinear & 570.59 & 2473 & 1.44 & 7.38 & 92.88\\
        CARAFE & +0.35 & +2 & 1.44 & 7.38 & 92.93 \\
        DySample & +0.10  & +0.3 & 1.44 & 7.38 & 92.93 \\
        FADE & +0.44 & +3 & 1.44 & 7.38 & 93.11 \\
        \bottomrule
        \end{tabular}
    \end{table}

    \begin{table}[!t]
        \footnotesize
        \centering
        \caption{Statistical Comparison of Model Performance}
        \label{tab:pvalue}
        \renewcommand{\arraystretch}{1.5}
        \begin{tabular}{@{}lcc@{}}
        \toprule
        \textbf{Parameter} & \textbf{\OurMethod} & \textbf{Fully Supervised} \\
        \midrule
        Experimental Data & $(93.47, 93.43, 93.47)$ & $(93.02, 93.01, 93.07)$ \\
        Mean$\pm$Std & $93.457\pm 0.023$ & $93.033\pm 0.032$ \\
        Variance & 0.0005 & 0.0010 \\
        \midrule
        $95\%$ CI & \multicolumn{2}{c}{$(0.362, 0.485)$} \\
        t-statistic & \multicolumn{2}{c}{$19.259$} \\
        P-value & \multicolumn{2}{c}{ $0.0001$} \\
        \bottomrule
        \end{tabular}
    \end{table}

    \subsection{Benchmark Comparison Experiments}\label{sec:Benchmark}
    We first present the experimental results by comparing \OurMethod with fully supervised baselines, semi-supervised approaches, and state-of-the-art general-purpose visual foundation models. The quantitative results are summarized in Table~\ref{tab:benchmark_03}.

    For \OurMethod, we evaluate the cumulative impact of integrating each key component into the ViT-Adapter backbone. Specifically, we report performance after (1) using depth-informed pseudo masks, (2) applying two-stage self-training refinement, and (3) incorporating the FADE dynamic upsampling module. The first row in Table~\ref{tab:benchmark_03} shows the result of a semi-supervised baseline using expanded data with conventional semi-supervised pseudo masks. We also include the performance of a fully supervised benchmark model trained only on $11,406$ manually annotated images, as well as two leading class-agnostic visual segmentation models: SAM and HQ-SAM. To this end, we incorporate the winning solution~\cite{cao2025first} of the Global Wheat Full Semantic Segmentation competition (GWFSS~\cite{wang2025global}) to represent a domain-specific approach.

    As shown in Table~\ref{tab:benchmark_03}, the semi-supervised approach achieves a mIoU of $92.36\%$. Incorporating only depth-informed pseudo masks raises the mIoU to $92.86\%$, while adding two-stage self-training slightly improves it to $92.88\%$. The gains from two-stage self-training appear limited on our dataset, primarily because performance has nearly reached saturation. This policy demonstrates more pronounced improvements in the DepthCropSeg experiments. Introducing the FADE module further increases the mIoU to $93.11\%$. The fully supervised benchmark model reaches an mIoU of $92.75\%$. In comparison, SAM and HQ-SAM achieve mIoUs of $32.80\%$ and $44.54\%$, respectively, on the test set of $6,760$ images. {The domain-specific approach reaches an mIoU of $69.50\%$.} We also conducte multiple experiments simultaneously on full supervision and \OurMethod under identical experimental settings. The results are shown in Table~\ref{tab:pvalue}. Statistical analysis indicates that the p-value is less than 0.001, demonstrating a significant difference.

    To further investigate the impact of upsampling operators on model performance, we compared various upsampling operators. As shown in Table~\ref{tab:compare_upsample}, comparative experiments reveal that all evaluated upsampling methods achieve competitive performance with only marginal increases in model complexity. Notably, FADE attains the highest mIoU of $93.11\%$, surpassing the bilinear baseline by $0.23\%$, with a negligible parameter increment of $0.44$M.
    
    We further evaluate SAM, HQ-SAM, and \OurMethod~on visual segmentation tasks with point-based prompts. For SAM and HQ-SAM, we randomly select 1–10 foreground and background points per image and report the best mIoU result per image. 

    Fig.~\ref{fig:SAM_compare} shows qualitative prediction examples from SAM, HQ-SAM, and \OurMethod~on representative crop images. The green pentagrams indicate foreground prompts, and red dots indicate background prompts.

    \begin{figure*}
        \centering
        \includegraphics[width=\linewidth]{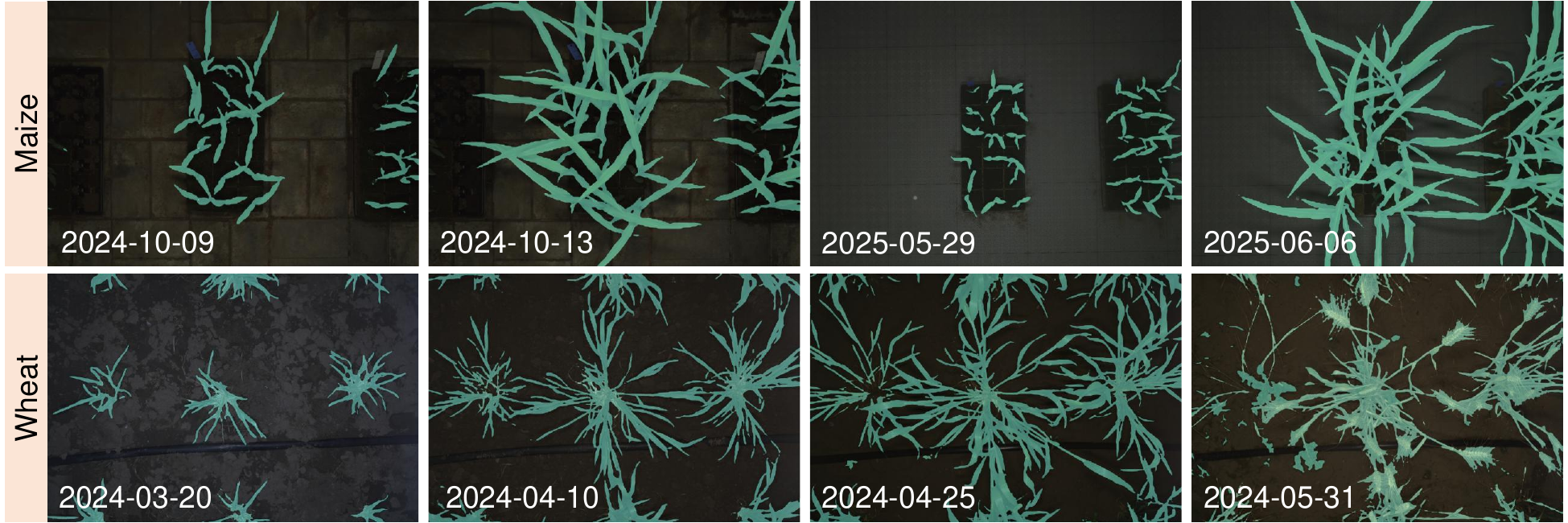}
        \caption{Segmentation Results of \OurMethod on Different Crops and Time Points}
        \label{fig:supp_infer}
    \end{figure*}

    \newcolumntype{E}{>{\centering\arraybackslash}p{1cm}}
    \begin{table*}[!t]\small
        \centering
        \begin{threeparttable} 
    	\caption{Performance (\%) Comparison on Specific Subsets. Best Performance Is in Boldface.}
    	\label{tab:SubData_performance}
            \renewcommand{\arraystretch}{1.25} 
            \addtolength{\tabcolsep}{0pt} 
           \begin{tabular}{@{}lEEEEEE@{}}
    			\toprule
    			\multirow{2}{*}{\textbf{Method}} & \multicolumn{2}{c}{\textbf{Soybean}} & \multicolumn{2}{c}{\textbf{Rice}}  & \multicolumn{2}{c}{\textbf{Fully Coverage}} \\ \cmidrule(lr){2-3} \cmidrule(lr){4-5} \cmidrule(lr){6-7}
    			& {mIoU} & {bIoU} & {mIoU} & {bIoU} & {mIoU} & {bIoU} \\ \midrule
    			
                    SAM   & 29.98 & 1.16 & 49.04 & 3.84 & 72.32 & 12.89 \\
                    HQ-SAM   & 35.15 & 4.17 & 69.12 & 21.41 & 78.97 & 15.83 \\
                    GWFSS Top-Solution   & 42.35 & 12.28 & 84.30 & 18.04 & 36.01 & 9.00 \\
                    DepthCropSeg   & 76.46 & \textbf{29.04} & 62.21 & \textbf{38.85} & 68.85 & 26.77 \\
    			DepthCropSeg++   & \textbf{90.09} & 17.38 & \textbf{86.90} & 21.35 & \textbf{99.86} & \textbf{47.42} \\
    			Fully Supervised   & 89.59 & 15.48 & 86.67 & 20.35 & 96.54 & 20.36 \\
    
    			 \bottomrule
    		\end{tabular}
        \end{threeparttable}
    \end{table*}

    \subsection{Performance Validation on Specific Datasets}
    To comprehensively evaluate the practical performance of the benchmark models and \OurMethod under different conditions, we conducted targeted experiments on three representative data subsets. Given the high cost of manual annotations in agricultural scenarios, most existing test sets are derived from the same sources as the fully supervised training data, which may introduce bias when assessing generalization. Therefore, we curated subsets that explicitly test the adaptability and robustness of our model:

    \begin{itemize}
        \item \textbf{Unseen soybean subset}: to evaluate adaptability to novel crop species not seen during training.
        \item \textbf{Nighttime rice subset}: to assess robustness under challenging imaging conditions with complex lighting and textures.
        \item \textbf{Full-coverage crop subset}: to examine performance in dense canopy scenes commonly encountered in field environments.
    \end{itemize}
    
    The quantitative results are summarized in Table~\ref{tab:SubData_performance}. In addition to mIoU, we report boundary Intersection over Union (bIoU), a metric designed to more accurately assess boundary segmentation performance.   

\begin{figure}[!t]
    \centering
    \includegraphics[width=0.95\linewidth]{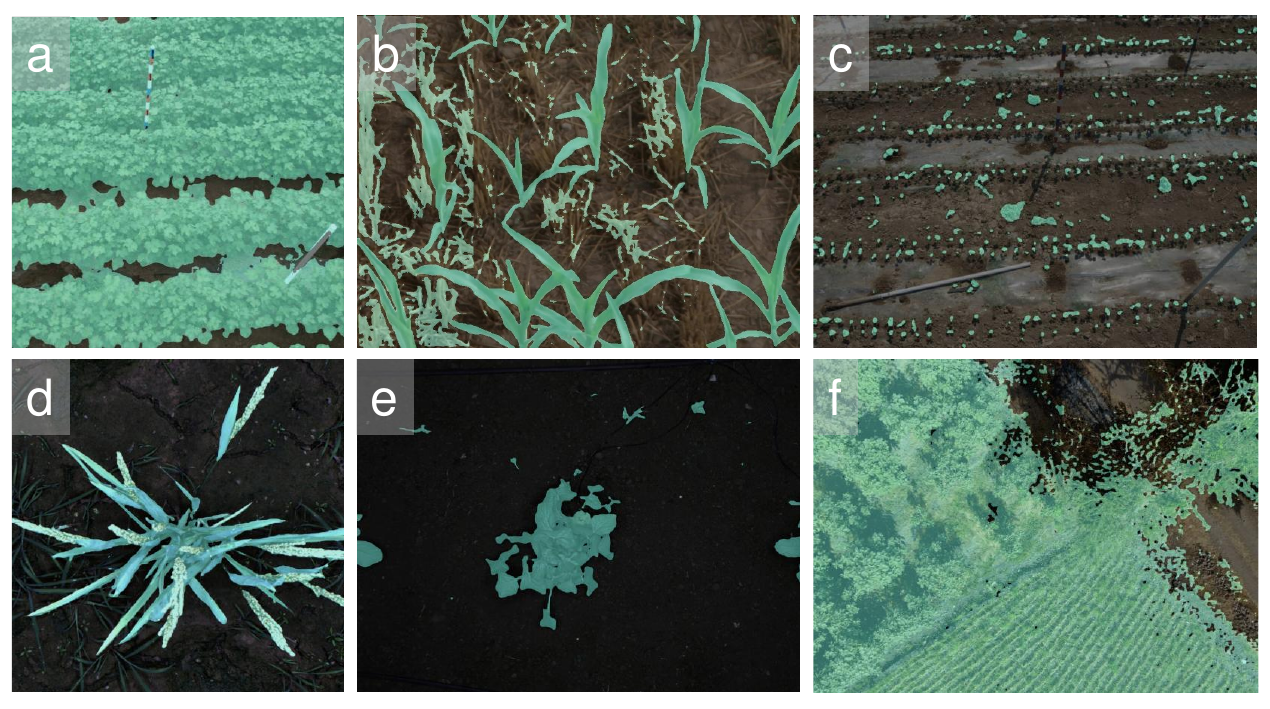}
    \caption{Some failure cases of \OurMethod.}
    \label{fig:failure_case}
\end{figure}
    
    On the unseen soybean subset, \OurMethod~achieves the highest mIoU of $90.09\%$, outperforming SAM ($29.98\%$), HQ-SAM ($35.15\%$), GWFSS Top-Solution ($42.35\%$), and the fully supervised model ($89.59\%$). Its bIoU also improves by approximately $1.9\%$ over the fully supervised model, demonstrating accurate boundary handling on unseen samples.

    On the nighttime rice subset, \OurMethod achieves an mIoU of $86.90\%$, the best among all models. 
    
    On the full-coverage crop subset, it reaches an almost perfect mIoU of $99.86\%$ and a bIoU of $47.42\%$. Notably, DepthCropSeg and GWFSS Top-Solution perform poorly on this subset (mIoU $68.85\%$ and $36.01\%$, respectively), reflecting its known failure cases in high-density scenes.

    Across all subsets, SAM and HQ-SAM perform significantly worse than agriculture-specific models, with low mIoU and bIoU scores, especially under challenging nighttime conditions.

    Furthermore, we attempted to evaluate our model on other crop datasets. Since most publicly available labeled data were already included in our training and testing sets, we conducted test visualizations using collected complex scene images. Specifically, we performed inference on images of maize and wheat crops at different growth stages under nighttime conditions. As shown in Fig~\ref{fig:supp_infer}, \OurMethod can handle crop recognition in such complex scenarios and has the ability to generalize across different growth stages.
    
    \begin{figure*}[!t]
        \centering
        \includegraphics[width=\textwidth]{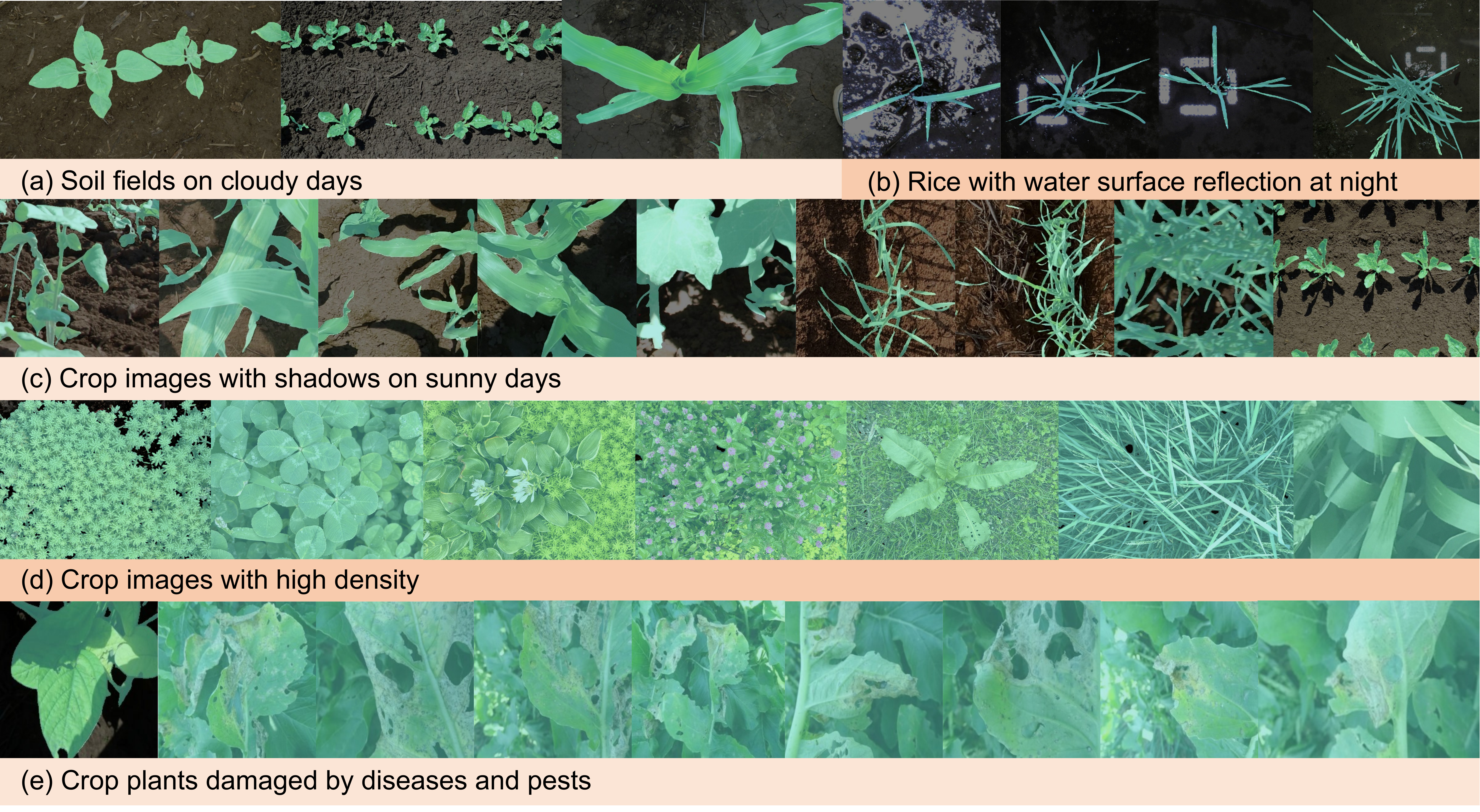} 
        \caption{Segmentation performance of \OurMethod under five representative agricultural field scenarios.}\label{fig:field_test}
    \end{figure*}
    
    \subsection{Qualitative Validation on Field Crop Data}
    \paragraph{Performance in Representative Field Scenarios}
    
    We evaluated segmentation performance across five representative agricultural field scenes:  (a) cloudy soil fields, (b) nighttime rice paddies with strong reflections, (c) sunny fields with high contrast and shadows, (d) densely planted crops, and (e) crops affected by pest or disease damage.  

    As shown in Fig.~\ref{fig:field_test}, \OurMethod achieves high-quality segmentation in all these challenging scenarios, producing smooth regional masks with clear boundaries. In particular, incorporating full-coverage samples in training significantly improves performance in dense canopy conditions. Additionally, the model effectively handles damaged or irregular plant structures commonly encountered in real fields.

\begin{figure*}[!t]
        \centering
        \includegraphics[width=\textwidth]{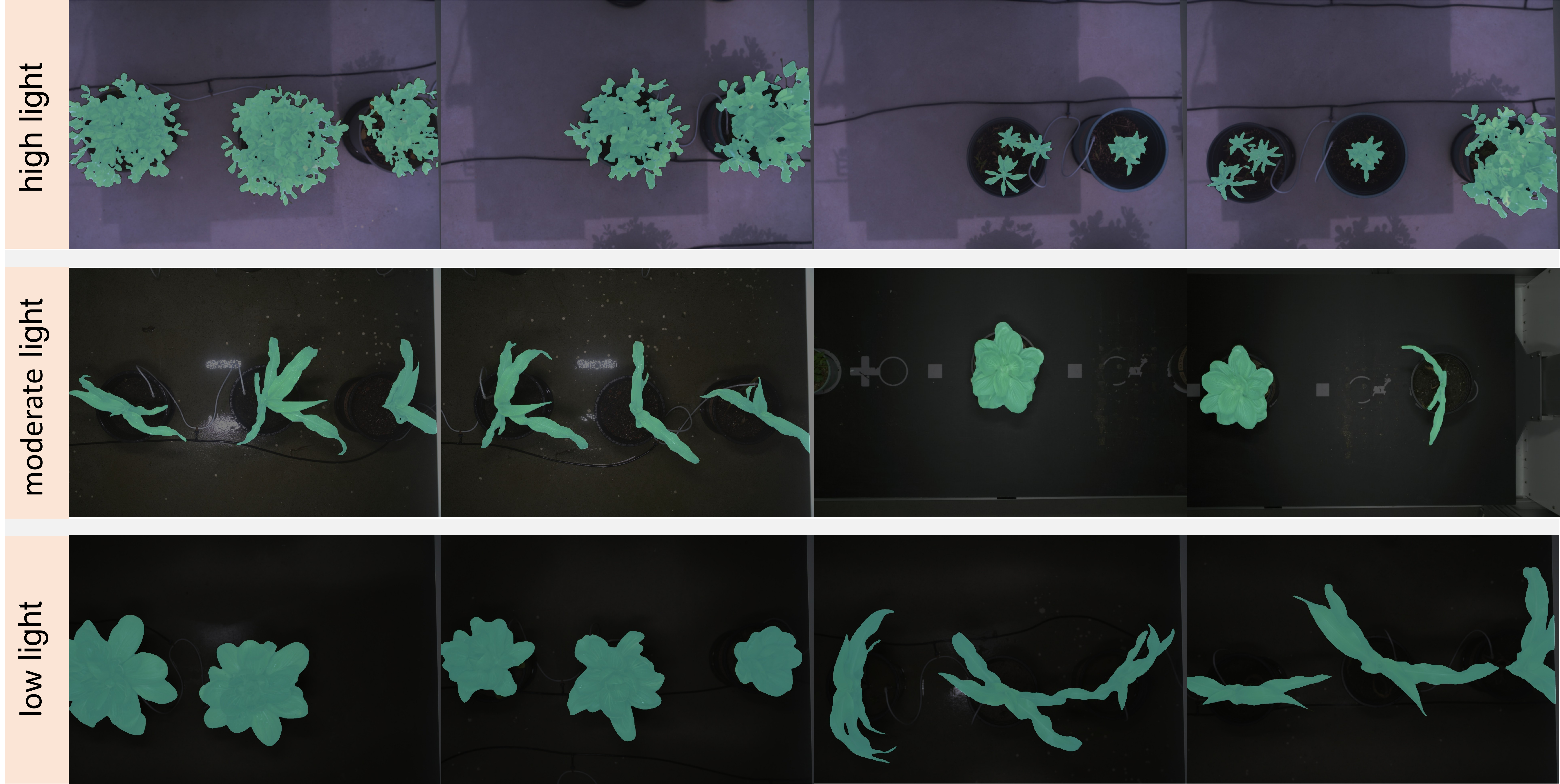} 
        \caption{Segmentation performance of \OurMethod~under varying illumination conditions.}\label{fig:illumination}
    \end{figure*}

    \paragraph{Robustness Under Varying Illumination}
    We evaluated OurMethod's robustness across illumination variations using crop images captured under strong, moderate, and low light conditions. As shown in Fig.~\ref{fig:illumination}, the model accurately segments crop regions even under low-light conditions where distinguishing the foreground from the background is challenging to the human eye. The segmentation quality remains stable across all illumination levels.
    
\begin{figure*}[!t]
        \centering
        \includegraphics[width=\textwidth]{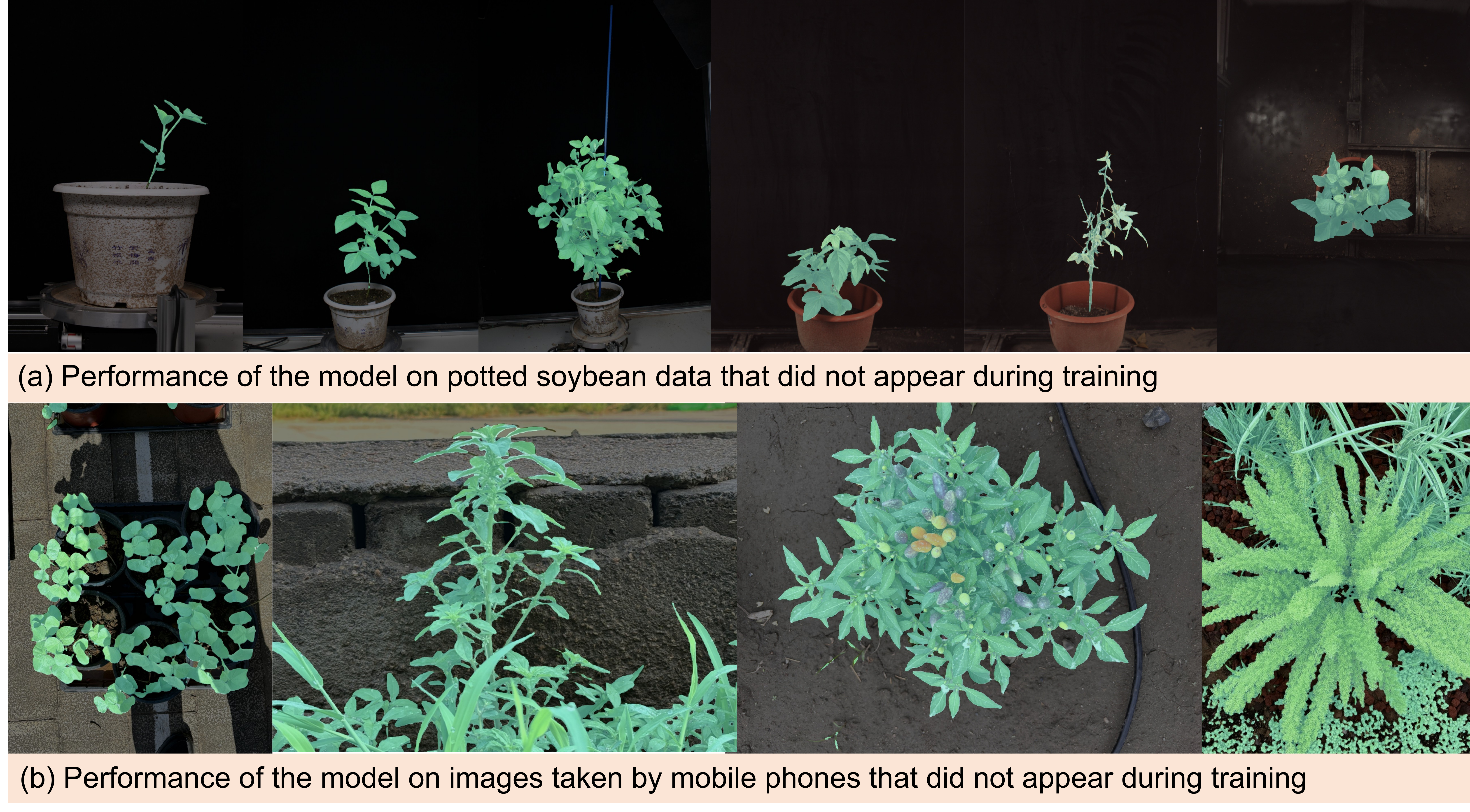} 
        \caption{Segmentation performance of \OurMethod~on unseen data}\label{fig:NeverSeen}
    \end{figure*}

    \paragraph{Adaptability to Crop Images Unseen During Training}
    Fig.~\ref{fig:NeverSeen} presents segmentation results on potted soybean plants that were never seen during training, including low-light conditions, different camera angles 
    (0\textdegree, 45\textdegree, 90\textdegree), and varying growth stages. The model consistently produces complete segmentations, accurately distinguishing foreground crops from background structures such as support rods. Additionally, we tested smartphone-captured images of ornamental plants that were also not present in the training set. Despite unfamiliar species and device characteristics, the model effectively excluded background shadows and irrigation pipes, accurately  identifying crop regions. These results demonstrate the model’s adaptability to crop images unseen during training, indicating its capacity to generalize beyond the specific crops and conditions seen in the training data.

    \subsection{Failure Cases}\label{sec:failure_case}
    Here we show some failure cases of \OurMethod. Although \OurMethod performs robustly across a wide range of species, scenes, and imaging conditions, several failure patterns remain observable in particularly demanding cases (Fig.~\ref{fig:failure_case}). Specifically, in high vegetation coverage scenarios (a), poles may be mis-segmented as vegetation. In scenes with withered weeds (b), a portion of the dried plants can be incorrectly classified as healthy vegetation. Repetitive row-wise soil patterns (c) occasionally introduce faint linear artifacts, while plants with certain complex morphologies (d) may be partially fragmented in the segmentation output. Under extremely low-light conditions (e), some soil clumps can be erroneously segmented as vegetation. In large-scale aerial views (f), extreme scale compression may lead to discontinuous segmentation.

    These cases reflect the inherent long-tail complexity of open-field imagery—where lighting, scale, and structural variation can deviate sharply from dominant patterns, even in substantially expanded datasets. They highlight potential directions for future improvement, such as enhancing structural cue extraction or developing adaptive scale modeling to further strengthen robustness.

\section{Discussion}
\label{sec:discussion}

\paragraph{DepthCropSeg Provides High-Quality Pseudo Labels for Baseline Training}
DepthCropSeg leverages Depth Anything V2 to generate monocular depth maps, from which pseudo crop segmentation masks are automatically derived based on depth features. This approach dramatically reduces the reliance on manual annotations. In our experiments, DepthCropSeg required only approximately $12$ hours to screen and obtain $12,927$ high-quality pseudo-labeled samples from an initial pool of $260,260$ crop images. 
By comparison, achieving comparable annotation quality and quantity through manual labeling would require a minimum effort of six months.
The experimental results demonstrate that these pseudo masks achieve sufficient accuracy for training: models trained on DepthCropSeg-generated pseudo labels surpassed those trained using conventional semi-supervised methods, achieving a $0.75\%$ improvement in mIoU (Table~\ref{tab:benchmark_03}). These findings indicate that DepthCropSeg offers an efficient and precise pseudo-labeling pipeline, laying a solid data foundation for training baseline models.

\paragraph{Key Optimization Strategies Improve Segmentation Performance}
To further enhance segmentation accuracy and boundary quality, we introduced two core optimization strategies into \OurMethod: a two-stage self-training framework and the FADE dynamic upsampling module.

First, the two-stage self-training approach refines the initial pseudo masks produced by DepthCropSeg through a confidence-guided filtering process. This strategy effectively mitigates the impact of noisy labels and improves the model’s ability to accurately recognize true crop regions. Although the overall improvement in mIoU is relatively modest, the contribution to model robustness is notable.

Second, replacing bilinear interpolation with the FADE dynamic upsampling module significantly improves the model’s boundary segmentation capabilities. Unlike traditional upsampling methods, FADE performs structure-aware dynamic kernel generation guided by encoder features, effectively reducing boundary blurring and detail loss. Incorporating FADE results in an approximate $0.25\%$ increase in mIoU and brings even more substantial improvements in edge precision, underscoring its importance to the overall performance of \OurMethod.

\paragraph{DepthCropSeg++ Demonstrates Outstanding Generalization}
\OurMethod exhibits strong segmentation performance across diverse agricultural scenes and crop types. By expanding the training dataset with large-scale, diverse, and complex scene data, including densely covered crop images paired with full-foreground pseudo masks, the model achieves high robustness under various conditions.
    
(1) On the full-coverage dataset, DepthCropSeg++ achieves a nearly perfect mIoU ($99.86\%$) and shows improvements of $3.32\%$ in mIoU and $27.06\%$ in bIoU compared to the fully supervised benchmark (Table~\ref{tab:SubData_performance}), demonstrating that incorporating dense canopy samples significantly enhances the model performance in this common field scenario.

(2) Across various real-world field conditions, such as different weather, plant densities, occlusions, and crops affected by pests or diseases (Fig.~\ref{fig:field_test}), the model maintains consistently high segmentation accuracy.

(3) Even in challenging low-light conditions, where human observers can barely distinguish crops from the background, the model delivers segmentation results comparable to those under high illumination (Fig.~\ref{fig:illumination}), indicating strong illumination robustness.

(4) \OurMethod also shows excellent adaptability to unseen crop types during training. On the nighttime soybean dataset, which was never seen during training, the model achieves an mIoU of $90.09\%$ (Table~\ref{tab:SubData_performance}). Furthermore, on potted soybean images with varying lighting, camera angles, and growth stages, none of which appeared in the training or test sets, the model produces accurate segmentations. Even on smartphone-captured ornamental plant images, the model effectively excludes background interference (Fig.~\ref{fig:NeverSeen}). These results suggest that \OurMethod~has learned cross-species generalizable features rather than relying solely on specific crop morphologies.

\section{Conclusion}
\label{sec:conclu}
In this work, we present \OurMethod, a foundational crop segmentation model tailored for complex agricultural scenarios. Trained on a large-scale, cross-scenario agricultural dataset with DepthCropSeg-generated pseudo-masks, \OurMethod reduces annotation costs and improves segmentation accuracy through a two-stage self-training strategy. It leverages ViT-Adapter for robust multi-scale feature extraction and integrates the FADE upsampling module to enhance boundary precision. Experimental results demonstrate its superior generalization and performance ($93.11\%$ mIoU), surpassing both fully supervised and general vision models. This study offers a new paradigm for efficient, high-performance agricultural perception, with future work aimed at 3D phenotyping and edge deployment.

\bibliographystyle{IEEEtran}
\bibliography{ref_depthcropseg++}

\end{document}